\pdfoutput=1

\documentclass[11pt]{article}

\usepackage{emnlp2021}

\usepackage{times}
\usepackage{latexsym}
\usepackage{amsmath}
\usepackage{amssymb}
\usepackage{array}
\usepackage{graphicx}
\usepackage{subfigure}
\usepackage{multirow}
\usepackage[encapsulated]{CJK}
\usepackage{float}
\usepackage{mathrsfs}
\usepackage{mathtools}
\usepackage{amsopn}
\usepackage{url}
\usepackage{soul}
\usepackage{longtable}
\usepackage{arydshln}
\usepackage{pgfplots}
\pgfplotsset{compat=1.15}

\usepackage{enumerate}
\usepackage{url}
\usepackage[T1]{fontenc}

\usepackage[utf8]{inputenc}

\usepackage{microtype}
\usepackage[draft]{changes}
\definechangesauthor[color=blue]{khc}
\title{Better Sign Language Translation with Monolingual Data}

\author{Ru Peng\textsuperscript{1$\ast$}, Yawen Zeng\textsuperscript{2}, Junbo Zhao\textsuperscript{1} \\
  \textsuperscript{1}Zhejiang University, Zhejiang, China \\
  \textsuperscript{2}ByteDance AI Lab, Beijing, China \\
  \texttt{rupeng@zju.edu.cn, yawenzeng11@gmail.com, j.zhao@zju.edu.cn}}

\date{}
\usepackage{multirow}
\usepackage{arydshln}
\usepackage{enumitem}
\usepackage{graphicx}
\usepackage{color}
\usepackage{footnote}
\makesavenoteenv{table}

\begin{document}
\maketitle
\begin{abstract}
Sign language translation (SLT) systems, which are often decomposed into video-to-gloss (V2G) recognition and gloss-to-text (G2T) translation through the pivot gloss, heavily relies on the availability of large-scale parallel G2T pairs.
However, the manual annotation of pivot gloss, which is a sequence of transcribed written-language words in the order in which they are signed, further exacerbates the scarcity of data for SLT.
To address this issue, this paper proposes a simple and efficient rule transformation method to transcribe the large-scale target monolingual data into its pseudo glosses automatically for enhancing the SLT translation.
Empirical results show that the proposed approach can significantly improve the performance of SLT, especially achieving state-of-the-art results on two SLT benchmark datasets PHEONIX-WEATHER 2014T and ASLG-PC12 \footnote[1]{Our code has been released at: \url{https://github.com/pengr/Mono\_SLT}}.
\end{abstract}

\section{Introduction}
\vspace{-4pt}
{Sign language translation (SLT) systems are to translate sign language video into target language with equivalent meaning automatically, thereby well assisting the communication between deaf community and society \cite{souza2017main}.
SLT is often decomposed into video-to-gloss (V2G) recognition and gloss-to-text (G2T) translation.
V2G system \cite{ZhouZZL20,Orbay9320278} first recognizes a sequence of glosses from sign language videos and then G2T system translates the recognized sign language glosses into written language.
In particular, the pivot gloss, which is a sequence of transcribed written-language words in the order in which they are signed, greatly reduces the difficulty of SLT, thereby achieving impressive results on SLT tasks \cite{schmidt2013using,app9132683,Camgoz9156773}.
\begin{figure}[!t]
\setlength{\belowcaptionskip}{-20pt}
\centering
\includegraphics[width=0.44\textwidth]{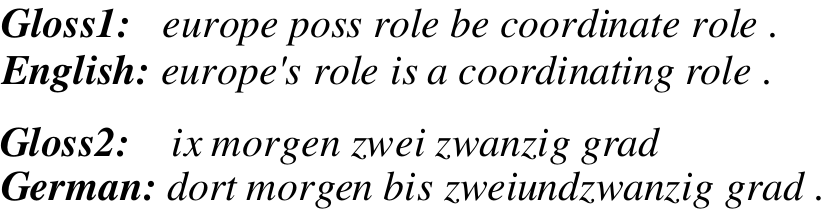}
\label{fig1}
\hfil
\centering
\caption{Two examples of Gloss-to-Text sentence pair.}
\label{fig:Fig1}
\end{figure}

Different from the written text bilingual machine translation scenario \cite{Bahdanau2015NeuralMT,vaswani2017attention}, SLT heavily relies on the availability of large-scale parallel G2T pairs. 
In other words, SLT is an low-resource machine translation scenario while is further exacerbated by the additional manual annotation cost of pivot gloss. 
The gloss annotation is different from its written counterpart, such as 
word expansion, morphology, deletions of stopwords, and so on.
For example, Figure \ref{fig:Fig1} shows two G2T sentence pairs. English words \textquotesingle$s, is, coordinating$ are transcribed into lemma forms such as $poss, be, coordinate$. German word $zweiundzwanzig$ expands as $zwei, und, zwanzig$. All stopwords $a, und, bis$ are omitted.  As a result, these differences will poses greater challenges of data scarcity for SLT \cite{Morrissey2010,stein2012analysis,hanke2010dgs,schembri2013building}.

To address this issue, this paper proposes a simple and efficient rule transformation method to transcribe the large-scale target monolingual data into its pseudo glosses automatically for enhancing the SLT translation.
Specifically, two transformation rules are defined for English and German to transcribe monolingual data into their sign language glosses. 
Also, we use the large-scale pseudo G2T data to pretrain the G2T translation system, and then further train it on the small-scale ground-truth G2T dataset.
Empirical results show that the proposed approach can significantly improve the performance of SLT, especially achieving state-of-the-art results on two SLT benchmark datasets PHEONIX-WEATHER 2014T and ASLG-PC12.
}

\section{Monolingual Data Augmentation}
\vspace{-5pt}
\textbf{Notations} 
Let $X$ and $Y$ denote gloss and text languages. Let $\mathcal{D}=\left\{ \left( {{\mathbf{x}}_{n}},{{\mathbf{y}}_{n}} \right) \right\}_{n=1}^{N}$ denotes the ground-truth G2T parallel corpus, where $N$ is the number of sentence pairs. Let $\mathcal{M}=\left\{ {{y}_{i}} \right\}_{i=1}^{I}$ denotes the monolingual text corpus, where $I$ is the number of monolingual sentences.

\subsection{Generating Sign language data}

\begin{table}[t]
\centering
\setlength{\abovecaptionskip}{0.cm}
\setlength{\belowcaptionskip}{-0.6cm}
\caption{Edit distance statistics of two benchmark SLT datasets.
\footnote[2]{WER denotes the word error rate. INS./DEL./SUB denote the number of insertions/deletions/substitutions, respectively. Tokens denote the number of gloss and text tokens.}}\label{table_I}
\resizebox{0.5\textwidth}{!}{%
\setlength{\tabcolsep}{0.6mm}{
\begin{tabular}{lllllll}
\hline
\hline
& WER & INS. & DEL. & SUB. & Gloss/Text Tokens\\
\hline
\textbf{Gloss-DE} & 86.4$\%$ & 797 & 39193 & 51780 & 75.7K/113.7K\\
\hline
\textbf{Gloss-EN} & 29.8$\%$ & 2656 & 116720 & 205141 & 910K/1030K\\
\hline
\hline 		
\end{tabular}}
}
\vspace{-10pt}
\end{table}


\textbf{Transformation Rules}
We analyze the differences between gloss and text of two benchmark SLT datasets as shown in Table \ref{table_I}. Compared with the insertions, deletions (removing stopwords and special chars, etc) and substitutions (morphological changes) are the primary behaviors of G2T transformation. To this end, we design two transformation rules to transcribe written text into sign language gloss for German and English SLT tasks as shown in Table \ref{table_II}. To easily understand for readers, we give the corresponding gloss transformed from the English and German sentences in Figure \ref{fig:Fig1}:

\textit{\textbf{EN-Gloss}: europe poss role be coordinate role .}

\textit{\textbf{DE-Gloss}: morgen bis zwei zwanzig grad}

We select the best transformation rules based on the BLEU scores between the ground-truth gloss and pseudo gloss generated by different sign language generation methods on the ground-truth parallel G2T dataset. The results are shown in Table \ref{table_III}. The optimal gloss generation quality of complete transformation rule is the reason we chose it, and removing any single rule will impair this quality. 

The quality of training sign language data generated by the back-translation model was superior to that generated by the transformation rule, while stands in entire contrast on the test data. This over-fitting issue directly led to the gap in the subsequent SLT performance between the two methods. Furthermore, both back-translation and transformation rules did not generate well enough German sign language data, which also revealed the reason for the poor performance of subsequent Gloss-DE task.
\begin{table}[!t]
\setlength{\abovecaptionskip}{0.cm}
\setlength{\belowcaptionskip}{-1.cm}
\centering
\caption{Transformation rules of Text-to-Gloss}
\label{table_II}
\centering
\setlength{\tabcolsep}{0mm}{\begin{tabular}{l|l}
\hline
\hline
\multicolumn{1}{c|}{\textbf{DE-Gloss}} & \multicolumn{1}{c}{\textbf{EN-Gloss}}\\
\hline
Omit punctuation & Clause segmentation\\
\hline
Multi-word token expansion & Abbreviation reduction\\
\hline
Omit and replace stopwords & Handling special chars\\
\hline
\multicolumn{1}{c|}{-} & Name entity recognition\\
\hline
\multicolumn{1}{c|}{-} & Lemmatization\\
\hline
\multicolumn{1}{c|}{-} & Omit function words\\
\hline
\hline 						
\end{tabular}}
\vspace{-4pt}
\end{table}

\begin{table}[!htbp]
\setlength{\abovecaptionskip}{0.cm}
\centering
\caption{BLEU score between the generated pseudo gloss and ground-truth gloss, where rule\{$i$\} denote the transformation rules that the $i$-th rule was removed.}
\label{table_III}
\centering
\setlength{\tabcolsep}{1mm}{\begin{tabular}{c|ccc|ccc}
\hline
\hline
\multicolumn{1}{c}{\multirow{2}*{}} & \multicolumn{3}{|c}{\textbf{DE-Gloss}} & \multicolumn{3}{|c}{\textbf{EN-Gloss}} \\
\cline{2-7}  
& Train & Dev & Test & Train & Dev & Test \\
\hline
\hline
bt & 14.80 & 11.04 & 9.51 & 98.39 & 96.22 & 96.34\\
rule & 7.03 & 13.41 & 11.54 & 96.20 & 96.34 & 96.75\\
rule1 & 7.03 & 13.41 & 11.54 & 95.90 & 96.07 & 96.42\\
rule2 & 5.31 & 11.41 & 9.28 & 94.87 & 94.83 & 95.62\\
rule3 & 1.94 & 2.31 & 1.93 & 96.13 & 96.23 & 96.70\\
rule4 & - & - & - & 95.88 & 96.04 & 95.99\\
rule5 & - & - & - & 57.78 & 58.01 & 57.52\\
rule6 & - & - & - & 67.69 & 67.06 & 66.79\\
\hline
\hline 						
\end{tabular}}
\vspace{-10pt}
\end{table}

\textbf{Back-Translation} we train the back-translation model ${{g}_{b}}$ : $Y\to X$ on the given parallel corpus $\mathcal{D}$, and then the model ${{g}_{b}}$ is used to decode the monolingual corpus $M$ to construct the pseudo G2T corpus ${{\mathcal{D}}_{p}}$.

\subsection{Data Augmentation}
\textbf{Combined Training} We train the translation model ${{g}_{c}}$ : $X\to Y$ on the combination of ground-truth and pseudo G2T corpora to decode the sign glosses.

\textbf{Further Training} We pre-train the translation model ${{g}_{f}}$ : $X\to Y$ on the pseudo G2T corpus ${{\mathcal{D}}_{p}}$, and further train it on the ground-truth G2T corpus $\mathcal{D}$ for fine-tuning.

\section{Experiments}
\subsection{Setup}
\textbf{Dataset} 
We evaluated the proposed method on PHOENIX-Weather 2014T \cite{Camgoz8578910} and ASLG-PC12 \cite{othman2012english} two benchmark SLT datasets. The training/validation/test set of PHOENIX-Weather 2014T contain 7096/519/642 sentence pairs, while that of ASLG-PC12 contain 82710/4000/1000 sentence pairs.
We extracted English and German monolingual texts from the mixed dataset of WMT14 and IWSLT15 English-German datasets for Gloss-to-English (Gloss-EN) and Gloss-to-German (Gloss-DE) SLT tasks. 
The mixed dataset consists of 3.97M sentence pairs with 94.37M English tokens and 90.90M German tokens.
To show the effect of monolingual data size on SLT performance, we randomly selected 1M sentences from the mixed dataset as a small-scale dataset for comparison.
We filtered out monolingual sentences that are longer than 60 to keep consistent with the SLT dataset, and followed by tokenization, no aggressive hyphen splitting, deescaped special chars.

\textbf{Setting} We used Stanza \cite{qi2020stanza} for text processing. English function words and German stopwords were collected by the vocabulary statistics of the SLT dataset. We truncated the vocabulary size of gloss and text languages to be 50K for both SLT tasks to cover the words in monolingual text corpus. Case-insensitive BLEU-4 \cite{papineni2002bleu} was used as the evaluation metric. We followed the experimental configurations of \cite{yin-read-2020-better} without using unknown replacing \cite{klein-etal-2017-opennmt}. For combined training, we trained 200K steps and 300K steps on small-scale and mixed datasets, respectively. We trained 100K steps for further training after that pre-trained 200K and 300K steps on small-scale and mixed datasets, respectively. All experiments were implemented by $fairseq$ toolkit \cite{ott-etal-2019-fairseq}.

\textbf{Baseline Systems}
We reported previous results on two benchmark SLT datasets, and reproduced the Transformer results of \cite{yin-read-2020-better} as a strong baseline. We also used monolingual data as gloss and text language to construct a pseudo SLT dataset for controlled experiment. Our systems were summarized as follows:
\vspace{-10pt}
\begin{itemize}[rightmargin=-2pt]
\setlength{\itemsep}{0pt}
\setlength{\parsep}{0pt}
\setlength{\parskip}{0pt}
\item[i.] +\textit{mono\_comb} $\&$ +\textit{mono\_fthr}: A controlled experiment by combined training or further training; 
\item[ii.] +\textit{bt\_comb} $\&$ +\textit{bt\_fthr} : The model was combined trained or further trained on the SLT dataset generated by back-translation.
\item[iii.] +\textit{rule\_comb} $\&$ \textit{+rule\_fthr}: The model was combined trained or further trained on the SLT dataset generated by transformation rules.
\end{itemize}

\subsection{Main Results}
Table \ref{table_IV} showed the results of all systems on the two benchmark SLT tasks. The $rule\_fthr$ achieved state-of-the-art SLT performance, which surpassed the baseline model 1.32 and 7.03 BLEU points on the Gloss-DE and Gloss-EN tasks, respectively. The results of controlled experiment was inferior to the baseline, which presented that it is necessary to generate more accurate sign language data. Since there existed a gap between the pseudo and ground-truth parallel SLT dataset, the performance of combined training was significantly lower than that of further training. Compared with back-translation, the model trained on the pseudo parallel SLT dataset generated by the transformation rules achieved considerable gains, which proved that our transformation rules were useful. The proposed methods existed obvious performance differences between the English and German SLT tasks, see section 3.6 for details.

\begin{table*}[!htbp]
\setlength{\abovecaptionskip}{0.cm}
\setlength{\belowcaptionskip}{-0.8cm}
\centering
\caption{G2T results on PHEONIX-WEATHER 2014T and ASLG-PC12 with mixed monolingual datasets.}
\label{table_IV}
\centering
\renewcommand{\arraystretch}{0.65} 
\setlength{\tabcolsep}{1mm}
{
\begin{tabular}{l|ccc|ccc}
\hline
\hline
\multicolumn{1}{c}{\multirow{2}*{Systems}} & \multicolumn{3}{|c|}{Gloss-DE} & \multicolumn{3}{c}{Gloss-EN}\\
\cline{2-7}  
& BLEU & ROUGE-L & METEOR & BLEU & ROUGE-L & METEOR \\
\hline
\multicolumn{7}{c}{\textit{Existing systems}}\\  
\hline
Raw data & 1.36 & 22.81 & 12.12 & 20.63 & 75.59 & 61.65\\
Preprocessed data & - & - & – & 38.37 & 83.28 & 79.06\\
RNN Seq2seq \cite{Camgoz8578910} & 19.26 & 45.45 & – & – & – & –\\
RNN Seq2seq \cite{Arvanitis9067871} & - & – & – & 65.9 & – & –\\
Transformer \cite{Camgoz9156773} & 24.54 & – & – & - & – & –\\
Transformer \cite{yin-read-2020-better}  & 23.32 & 46.58 & 44.85 & 82.41 & 95.87 & 96.46\\
Transformer Ens. \cite{yin-read-2020-better} & 24.90 & 48.51 & 46.24 & 82.87 & 96.22 & 96.60\\
\hline
\multicolumn{7}{c}{\textit{Our systems}}\\  
\hline
Transformer & 23.29 & 46.96 & 45.25 & 82.36 & 95.23 & 96.00\\
\quad+mono\_comb & 20.85 & 43.07 & 41.61 & 80.18 & 95.11 & 96.10\\
\quad+mono\_fthr & 21.83 & 44.84 & 43.13 & 77.92 & 94.33 & 95.18\\
\quad+bt\_comb & 19.54 & 41.37 & 40.00 & 88.15 & 96.51 & 96.65\\
\quad+bt\_fthr & 21.16 & 44.14 & 42.67 & 88.71 & 96.61 & 96.73\\
\quad+rule\_comb & 20.25 & 44.03 & 41.76 & 89.44 & 97.10 & 97.16\\
\quad+rule\_fthr & \textbf{24.64} & \textbf{48.70} & \textbf{47.04} & \textbf{89.90} & \textbf{97.19} & \textbf{97.29}\\
\hline
\hline 						
\end{tabular}}
\end{table*}

\begin{table*}[!htbp]
\setlength{\abovecaptionskip}{0.cm}
\setlength{\belowcaptionskip}{-0.2cm}
\centering
\caption{Results of ablation experiments with different transformation rules}
\label{table_V}
\centering
\begin{tabular}{l|ccccccc}
\hline
\hline
Systems & rule\_fthr & rule1 & rule2 & rule3 & rule4 & rule5 & rule6\\
\hline
\hline
Gloss-DE & 24.64 & 24.40 & 25.05 & 23.06 & - & - & -\\
Gloss-EN & 89.90 & 90.78 & 90.34 & 90.20 & 90.49 & 86.24 & 59.51\\
\hline
\hline 						
\end{tabular}
\end{table*}

\begin{table}[!t]
\setlength{\belowcaptionskip}{-0.4cm}
\centering
\caption{Time cost (in seconds) of generating sign language data. Bracket value denotes the time cost of generating English SLT data without NER and lemmatization.}
\label{table_VI}
\centering
{\begin{tabular}{c|ccc|ccc}
\hline
\hline
Method & Gloss-DE & Gloss-EN\\
\hline
\hline
bt & 20240 & 33386\\
rule & 168.6 & 38779 (416.7)\\
\hline
\hline 						
\end{tabular}}
\vspace{-15pt}
\end{table}

\subsection{Effect of monolingual data size}
\begin{figure}[!htbp]
\setlength{\abovecaptionskip}{0.cm}
\setlength{\belowcaptionskip}{-0.2cm}
\centering
\subfigure[Gloss-DE task]{\includegraphics[width=0.38\textwidth]{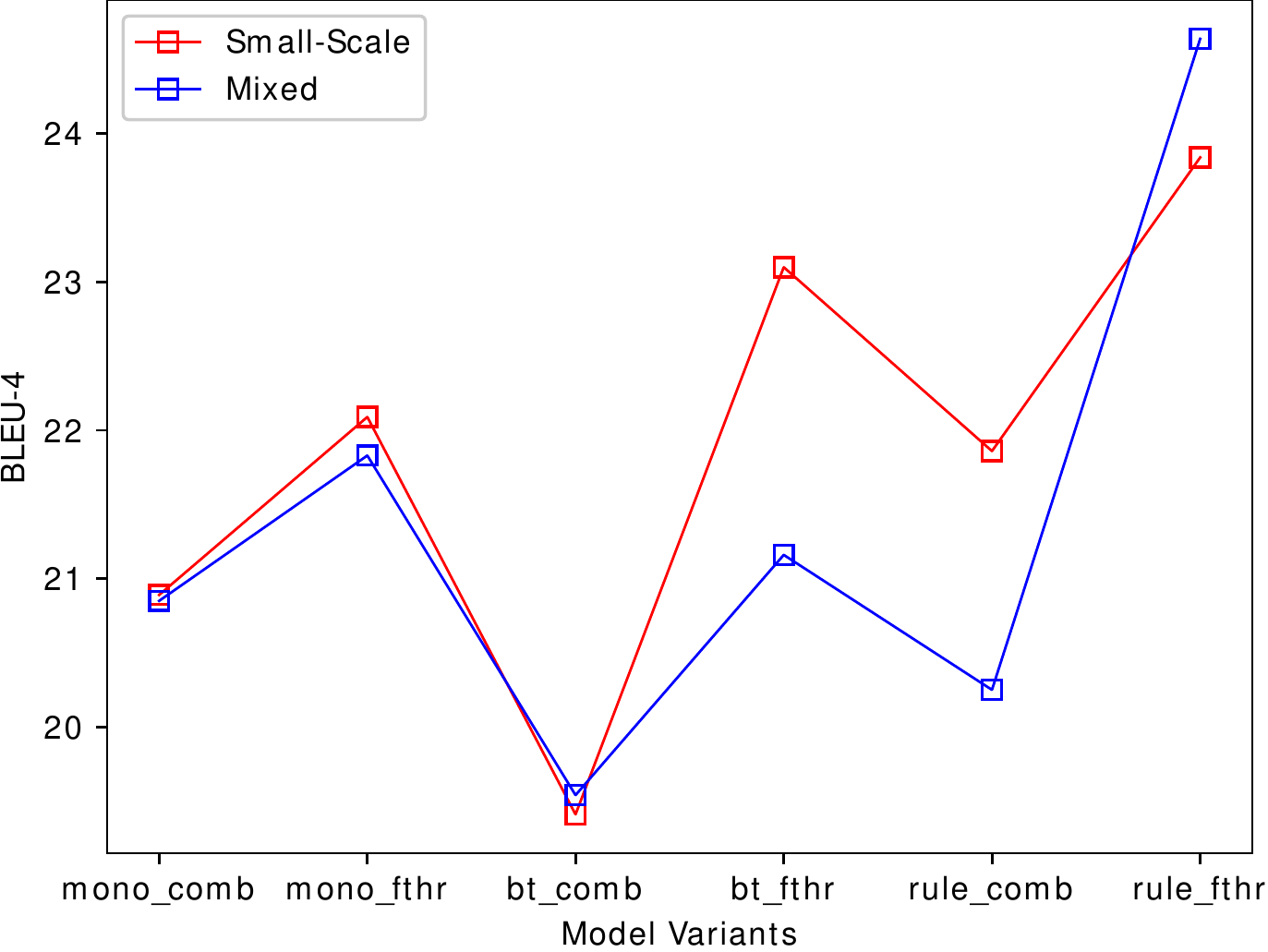}
\label{fig2_a}}
\hfil
\subfigure[Gloss-EN task]{\includegraphics[width=0.38\textwidth]{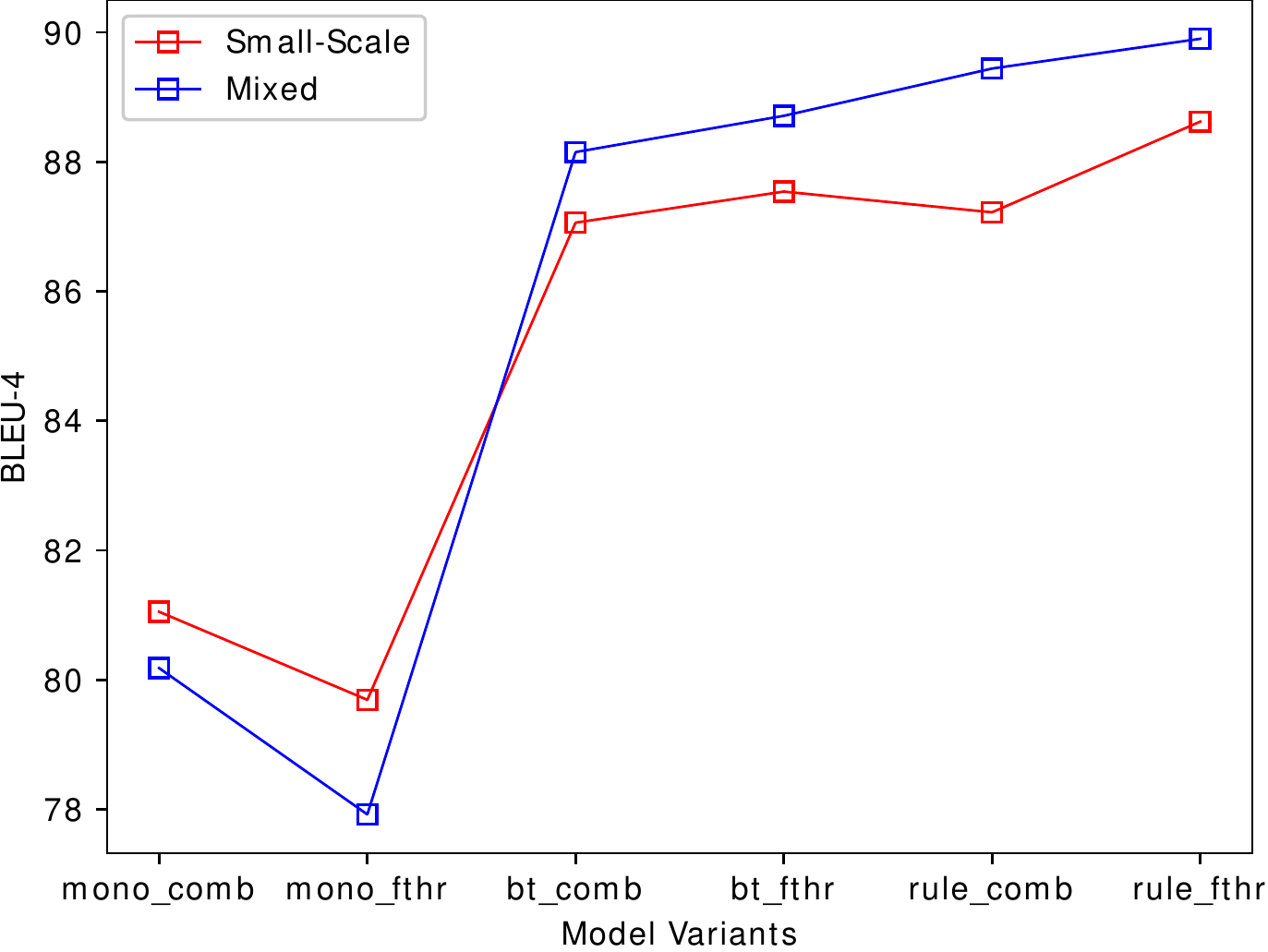}
\label{fig2_b}}
\hfil
\centering
\caption{Comparison of the proposed methods on small-scale and mixed monolingual datasets.}
\label{fig:Fig2}
\end{figure}

Figure \ref{fig:Fig2} showed the effect of different monolingual data sizes on the proposed method. On the Gloss-DE task, due to the non-ignorable gap between the pseudo and ground-truth German SLT datasets, the larger monolingual dataset would carry more noises to lead the translation performance descended. However, $rule\_fthr$ relied on independent rules and training to avoid overfitting noises and acquire substantial gains. The results on Gloss-EN task fully embodied that the growth of monolingual data size 
could bring a significant performance advancement to the proposed methods.

\subsection{Ablation Study}
We performed ablation experiments to evaluate the impact of each transformation rule on the final results as shown in Table \ref{table_V}. On the Gloss-DE task, removing multi-word token extension ($rule2$) would bring 0.41 BLEU points improvement to the single best model, even exceed 0.15 BLEU points for the ensemble model of Transformer. On the Gloss-EN task, removing any of the first 4 rules could improve the performance of single best model, and the best improvement was 0.88 BLEU points. The cliff-like decline of $rule6$ indicated that English sign language gloss needs to remove function words compared with text. In general, we should further explore more simpler rules to avoid overfitting.

\subsection{Efficiency of sign language generation}
Table \ref{table_VI} given the time cost of sign language data generated by back-translation and transformation rules on the mixed monolingual dataset.
The transformation rules were far more efficient than back-translation on the Gloss-DE task, while since the text processing using Stanza was slower than back-translation on the Gloss-EN task. 
It could be alleviated by using an efficient tool.

\subsection{English SLT vs. German SLT}
In this section, we explained why the proposed method failed on the Gloss-DE task. Firstly, it was more difficult for the model to learn the translation from German sign language gloss to text, which was presented in the results of Raw data row in Table \ref{table_III}. The second lied in the greater difference between German gloss and German text, as shown in Table \ref{table_VII}. Because German sign language was manually transcribed by experts without a fixed standard, while English sign language was constructed semi-automatically by a rule-based approach.
\begin{table}[!t]
\setlength{\belowcaptionskip}{-0.4cm}
\centering
\caption{Token similarity statistics (the proportion of the same word) between gloss and text.}
\label{table_VII}
\centering
\setlength{\tabcolsep}{1mm}{\begin{tabular}{lllll}
\hline
\hline
& Train & Dev & Test\\
\hline
Gloss-DE & 15.59$\%$ & 15.10$\%$ & 14.82$\%$\\
Gloss-EN & 98.23$\%$ & 98.21$\%$ & 98.23$\%$\\
\hline
\hline 						
\end{tabular}}
\vspace{-9pt}
\end{table}

\section{Conclusion}
In this paper, we proposed a simple and effective transformation rules and back-translation to transcribe monolingual text into its pseudo gloss data. 
We then exploited the pseudo parallel SLT dataset for data augmentation, and achieved state-of-the-art results on the two existing benchmark SLT tasks. 
Crucially, we conducted extensive experiments to analyse the effect of our method.

\bibliography{anthology}
\bibliographystyle{acl_natbib}

\end{document}